\documentclass[a4paper,fleqn]{cas-sc}

\usepackage[authoryear,longnamesfirst]{natbib}

\def\tsc#1{\csdef{#1}{\textsc{\lowercase{#1}}\xspace}}
\tsc{WGM}
\tsc{QE}
\tsc{EP}
\tsc{PMS}
\tsc{BEC}
\tsc{DE}

\begin{document}
\let\WriteBookmarks\relax
\def\floatpagepagefraction{1}
\def\textpagefraction{.001}
\shorttitle{A coordinated multi-layer runtime monitoring framework for AI systems}
\shortauthors{Niclas Flehmig et~al.}

\title [mode = title]{COMET: A coordinated multi-layer runtime monitoring framework for human-supervised AI-enabled safety systems}                      
\author[1]{Niclas Flehmig}[orcid=0009-0006-6458-2689]
\cormark[1]
\ead{niclas.flehmig@ntnu.no}

\credit{Writing -- review \& editing, Writing -- original draft, Methodology, Investigation, Formal analysis, Data curation, Validation, Visualization, Conceptualization}

\affiliation[1]{organization={Department of Mechanical and Industrial Engineering, Norwegian University of Science and Technology},
               city={Trondheim},
                country={Norway}}

\author[2]{Mary Ann Lundteigen}[orcid=0000-0002-9045-6815]

\credit{Writing -- review \& editing, Supervision}

\author[1]{Shen Yin}[orcid=0000-0002-3802-9269]

\credit{Writing -- review \& editing, Supervision}

\affiliation[2]{organization={Department of Cybernetics Engineering, Norwegian University of Science and Technology},
                city={Trondheim},
               country={Norway}}

\cortext[cor1]{Corresponding author}

\begin{abstract}
Incorporating AI systems into safety systems requires monitoring to avoid silent faults and ensure safe and reliable operation. Single monitors often do not provide actionable information for human supervisors, who act as the system's adaptive controller. We present COMET, a coordinated multi-layer runtime monitoring framework for human-supervised AI-enabled safety systems in which each monitor is anchored in a different fault mode. Monitor outputs are combined in a coordination layer and mapped by a communication layer to a traffic-light display. In this work, COMET consists of three monitoring layers: signal integrity, physics consistency, and the AI system's competence envelope. The framework is evaluated on an open dataset of compressed-air leaks, where an AI system detects leaks from acoustic emissions, with held-out conditions simulating covariate shift, semantic shift, and sensor faults, events expected during runtime but unknown in their exact appearance. All monitors are frozen after development and tested zero-shot, and the framework is compared against no-monitoring and single-monitor setups. The bare AI system fails silently and confidently under all three shift types, whereas COMET reduces the trusted-wrong-answer rate to zero on covariate-shifted no-leak data and on three of four sensor faults, at a nominal false-alarm rate of $\approx 1\%$. Single monitors come close in detection but cannot attribute the fault mode. COMET converts their anonymous detection flags into diagnoses. Residual silent misses under semantic shift define the boundary of passive monitoring. Runtime monitoring frameworks for AI-enabled safety systems must therefore combine quantitative fault detection and isolation with qualitative, human-oriented feedback.
\end{abstract}


\maketitle

\section{Introduction}
\subsection{Motivation}
The diffusion of AI systems from research into real-world applications has accelerated in recent years, driven by cheap compute, capable models, large amount of available data, and increasingly complex systems that need to be managed. Whereas for consumer products reliable operation of AI systems is largely a question of practicality, for safety-critical systems safe and reliable operation is a requirement, not a nice-to-have. This requirement is emphasized by policies and standards such as the EU AI Act \citep{euaiactEuropeanUnionArtificial2024} and the ISO/IEC TR 5469 \citep{iso/iectr5469:2024ArtificialIntelligenceFunctional2024}, which addresses the incorporation of AI systems into systems whose failures can cause severe consequences. One means of ensuring reliable and safe operation is monitoring \citep{rausandSystemReliabilityTheory2021, levesonEngineeringSaferWorld2012}. It provides information about system behavior during runtime and thus contribute to maintaining safe and reliable operation. Throughout this work, the term \textit{AI system} describes a system consisting of an AI model and its necessary pre- and post-processing steps during runtime, whereas \textit{AI-enabled system} describes the broader engineered system into which one or more AI systems are integrated \citep{detnorskeveritasAssuranceAIenabledSystems2023}.

In recent years, much research has focused on developing monitors for AI systems, most commonly are out-of-distribution detectors \citep{yangGeneralizedOutofDistributionDetection2024, luOutofDistributionDetectionTaskOriented2024}. These works typically aim to improve the performance on new or established benchmarks rather than proposing holistic frameworks, where not only the quantitative aspect of detecting samples outside the learned distribution matters, but also the qualitative aspects on how detections are communicated, and how monitors work together with the human supervisors who are supposed to use them. A few works have begun to combine multiple monitor into unifying frameworks \citep{darioUnifyingRuntimeMonitoring2027, torensRuntimeMonitoringOperational2025, ferreiraraulsenaRuntimeSafetyMonitoring2023}, but these still address detection rather than diagnosis and do not consider the human supervisor. This leads to the question:

\begin{quote}
    How can we design a runtime monitoring framework for AI systems that detects potentially unreliable system behavior, isolates faults, and supports the human supervisor in the loop, so as to enhance safe and reliable operation?
\end{quote}

The purpose of this work is to propose and test such a framework under realistic conditions. Derived from fault-tolerant control theory, where adapting the control of a system requires fault diagnosis, i.e., detection, isolation and identification, the framework promotes safe and reliable operation by detecting shifts that can degrade the AI system's performance and by providing qualitatively meaningful information to the human supervisor. To this end, we select a dataset and an AI system that together represent an AI safety system, and build COMET, a coordinated multi-layer runtime monitoring framework whose monitors cover different fault modes, are fused through a coordination layer, and are mapped to a traffic-light-based display through a communication layer. The framework is derived in Section~2. Section~3 describes the experimental setup, which tests the framework under realistic conditions and compares it with alternative monitoring strategies. The results are presented in Section~4 and discussed in Section~5. Section~6 concludes and outlines future work.

\subsection{Related work}
The need for monitoring deployed AI systems is already established in the MLOps literature \citep{kreuzbergerMachineLearningOperations2023}, where monitoring serves continuous performance improvement. However, its scope is operational rather than safety-related, i.e., the monitored systems are typically not part of safety functions, and reliability and safety implications of unreliable outputs are not the focus. On the regulatory side, the EU AI Act mandates post-market monitoring for high-risk AI systems but does not specify how such monitoring should be realized \citep{euaiactEuropeanUnionArtificial2024}. Similarly, ISO/IEC TR 5469 on AI and functional safety names monitoring as a potential risk-reduction measure without detailing its concrete design \citep{iso/iectr5469:2024ArtificialIntelligenceFunctional2024}. This work addresses that gap of a concrete runtime monitoring framework for AI-enabled safety systems, where system reliability is a primary concern and a human supervisor remains in the loop as highest control entity.

The most common monitoring technique is out-of-distribution (OOD) detection \citep{yangGeneralizedOutofDistributionDetection2024, luOutofDistributionDetectionTaskOriented2024}. In simple terms, it discriminates between samples inside the learned distribution, where the AI system should operate as trained and tested, and samples outside it, where outputs are potentially unreliable. These methods perform well on established benchmarks, with effectiveness depending on the method used. They can be applied using in-distribution data alone \citep{liu2023unsupervised, heng2024out} or a combination of in-distribution and OOD data \citep{liu2020energy}, and many are training-agnostic, operation post-hoc on a deployed model \citep{hendrycksBaselineDetectingMisclassified2016, sunOutofDistributionDetectionDeep2022}. In concrete applications, OOD detection has been used to increase trustworthiness of deployed AI systems, for instance in prognostics and health management for renewable energy systems \citep{xieUnifiedOutofdistributionDetection2023} and in mechanical fault diagnostic \citep{xiaoUncertaintyawareDeepVariational2025}. What these method have in common is that they are typically proposed and evaluated as single, isolated monitors, not as part of a holistic runtime monitoring framework.

Combining complementary monitors into holistic runtime monitoring frameworks that cover different fault modes and are potentially more robust than single monitors is underrepresented in current research. The few existing examples \citep{ferreiraraulsenaRuntimeSafetyMonitoring2023, darioUnifyingRuntimeMonitoring2027, torensRuntimeMonitoringOperational2025} demonstrate that combining monitors can improve detection, but they treat monitoring as a purely algorithmic output where the human supervisor who may need to perceive, interpret and decide on the outputs is not explicitly considered. Moreover, existing combinations fuse monitor to improve detection performance, rather than coordinating local diagnoses toward fault isolation in the sense of classical fault diagnosis. This work addresses both gaps. It develops a runtime monitoring framework with multiple monitor derived fault-tolerant system principles, and it designs the framework around the human in the loop.

Considering the human in the loop means going beyond bare in/out verdict of an OOD detector. Such a verdict may not be sufficient when a supervisor remains responsible and needs a clear basis for judgment. To take over from an AI system or override a potentially unreliable output, they must understand what kind of event is occurring. Otherwise, even well-intended monitoring can contribute to situations similar to automation surprise \citep{sarterTeamPlayPowerful1997}. Integrating a communication layer and building a more holistic runtime monitoring framework can thus promote more resilient, reliable, and safe system operation. This work address how outputs from multiple monitors can be fused into qualitatively richer feedback for humans in the loop.

\subsection{Contributions}
Based on the related work and the limitations identified above, this works makes the following contributions to the field of runtime monitoring for AI systems:

\begin{enumerate}
    \item We present COMET, a coordinated multi-layer runtime monitoring framework in which multiple monitors derive their authority from different fault modes. In this work, we have three monitors: engineering rules (signal integrity), first-principle physics (spectral leak template), and learned behavioral statistics (feature-space competence envelope). This results in fault modes that are disjoint by construction. The monitor outputs are combined by a coordination layer and fused into a fixed taxonomy with a traffic-light mapping, designed as the communication layer to the human supervisor in the loop.
    \item We demonstrate the quantitative performance via an empirical evaluation on an open dataset that serves as a proxy for a safety-critical application. Covariate shift, semantic shift, and injected sensor faults are held out from all training, validation, and calibration. Every monitor is frozen after development, and the shifts are evaluated exactly once, zero-shot. An ablation ladder compares COMET against the bare AI system and single-monitor setups, including the commonly used max-softmax probability (MSP) baseline, using supervisor-oriented metrics such as the trusted-wrong-answer rate and silent-miss-coverage.
    \item We derive design insights for runtime monitoring of deployed AI systems, supported by the ablations and by pre-registered sensitivity variants, i.e., why disjoint anchoring matters more than individual monitor strength, why the calibration unit must match the decision unit, and where the boundary of passive monitoring lies. We report the negative results and residual fault modes openly, including the silent misses no passive layer combination caught, which we consider essential evidence for future research on monitoring AI systems.
\end{enumerate}

\section{Methods}

The first part of the methods section describes the AI application that was used to represent an AI-enabled safety system, the complementary dataset, the specific model architecture, the pre-processing of the data, and the training process. Then, the coordinated multi-layer runtime monitoring framework is derived from the concept of fault-tolerant systems and coordinated fault diagnosis. The specific monitors that are used in this work are derived. Finally, the specific taxonomy states for the coordination and communication layer is described.

\subsection{AI application: leak detection system}\label{sec21}
To evaluate the coordinated multi-layer runtime monitoring framework, we used an open dataset \citep{johnsonIDMTISACompressedAirDataset2020} on compressed-air leaks with the complementary AI system that performs well on the dataset with accuracies reaching over 95 \% \citep{johnson2020compressed}. Its high performance reflects a realistic scenario where the result from training and testing satisfies the requirement for real-world deployment of an AI system. The AI system used was designed to classify spectrogram files into "leak" or "no leak" where a compressed-air leak represents a costly fault in industrial productions. The leak of compressed air does not represent a safety-critical issue, but using acoustic emissions together with AI systems to detect leaks is a common method used in process industries \citep{banjaraMachineLearningSupported2020, ullahPipelineLeakageDetection2023, xuGasLeakageIntelligent2026}. The detection of leaks in this domain represents a safety system that is essential to ensure safe and reliable operation. Due to the lack of open and diverse data on gas leaks but the similarity between the use cases, we chose the compressed-air leak detection as suitable proxy for the safety-critical leak detection in process industries.

In this work, the AI system represents the AI component inside an AI-enabled safety system. It performs the safety function of detecting the leak, whereas the overall system additionally comprises physical components such as the sensors, the monitored process, and potential actuators that are activated to maintain safe and reliable operation.

\subsubsection{Dataset}\label{sec211}
\begin{figure}
    \centering
    \includegraphics[width=0.8\linewidth]{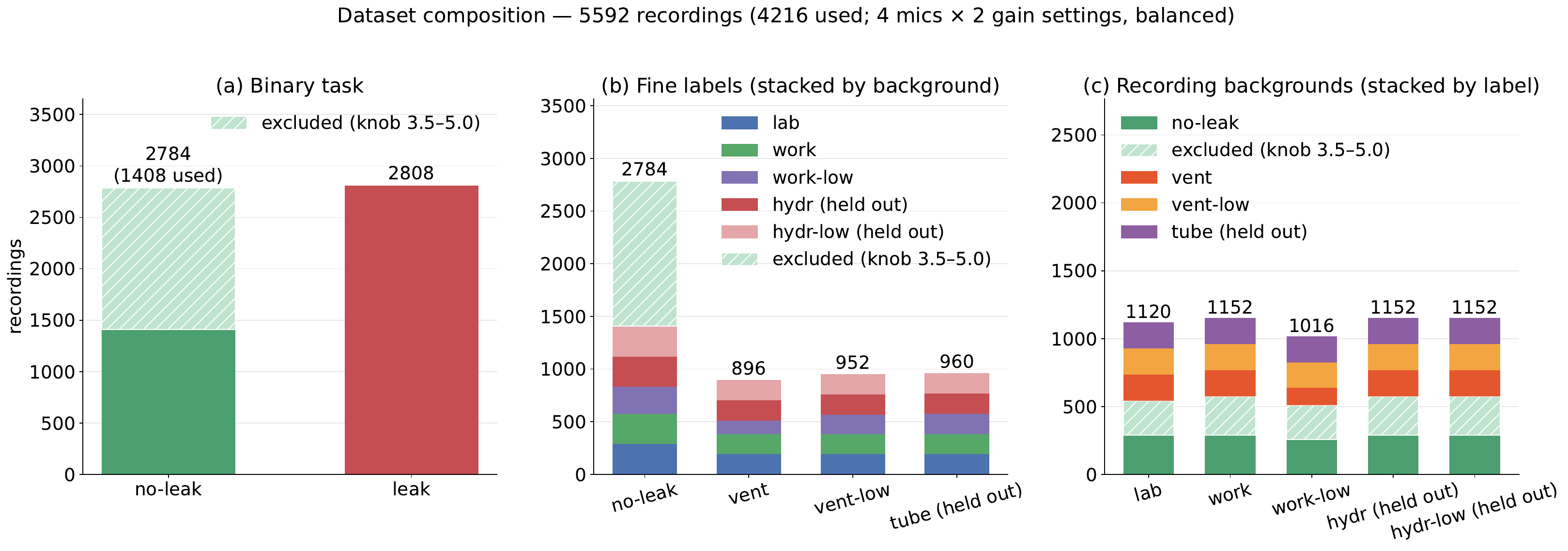}
    \caption{Distribution of the 5592 recordings (4 microphones × 2 gain settings) across the experimental labels. (a) Binary classification task: no-leak vs. leak recordings; the no-leak band at knob settings 3.5–5.0 (hatched) was excluded from all experiments to keep a clear severity margin between the classes, leaving 4216 usable recordings. (b) Fine-grained leak labels ("no-leak", "vent", "vent-low", "tube") stacked by recording background; tube leaks and both hydraulic ("hydr"/"hydr-low") backgrounds were never used for training or calibration and serve exclusively as the held-out semantic and covariate shift conditions. (c) The complementary view: recording backgrounds stacked by leak label, showing that every background contains a balanced mix of leak types.}
    \label{fig:dataset_composition}
\end{figure}

The data used for this work were collected by the Fraunhofer institute \citep{johnsonIDMTISACompressedAirDataset2020}. The data represent compressed-air leaks that were collected with 4 microphones, at 2 gain settings, with different backgrounds (i.e., lab noise "lab", workshop noise "work"/"work-low", hydraulic noise "hydr"/"hydr-low"), different leak types (i.e., vent leak at 6 bar "vent", vent leak at 5 bar "vent-low", tube leak "tube"), and different severity gradation based on the knob state. Each leak type contains all backgrounds and for each background there are three sessions of recording. There are two recordings for each knob state, one with low gain and one with max gain. The knob state increases from 0 to 3 with a step size of 1 and from there it continues with a step size of 0.5 until 9.0. The knob state represents the number of turns to open the vent where 0.5 represents half a turn. Between knob state 3.5 and 5.0 there are small leaks present but not detectable for the installed microphones thus these knob states are excluded from the data. The whole dataset composition is seen in Figure~\ref{fig:dataset_composition}.

\subsubsection{Primary AI system}
The primary AI system used for this work is based on the original convolutional neural network (CNN) model from \citet{johnson2020compressed} that was used on the same dataset with some adaptions. The model architecture consists of an initial block that applies a Gaussian noise with standard deviation of 0.1 and that is only active during training mode. Next, there are three convolution blocks with a GELU \citep{hendrycksGaussianErrorLinear2016} activation function and max pooling with channels 16, 32 and 64. The final block contains a dropout function, a dense multi-layer perceptron (MLP) network, a GELU activation function, another dropout regularization, and finally a second MLP with two logits as output. The whole model has 121,794 parameters.


\subsubsection{Pre-processing}\label{sec213}
Each recording has a length of 30 seconds and a measurement frequency of 48 kHz. First each recording was transformed with a short-time Fourier transformation (STFT) to create a spectrogram of shape 513 x ~2813 (frequency bins x time frames). Then the spectrogram is cut along the time axis into non-overlapping 24-frame patches, ~267 ms. One recording yields 117 patches, each of shape 513 x 24. This also is the fixed-size "image" for the AI system. Every patch inherits its recording's label: "leak" or "no-leak". Next, we augmented the training patches with noise to make the model more robust. This is relevant for the patches with lab noise which basically means no added noise. In the real world, we always have noise so this is why we chose the augmentation. With a probability of 0.5, we added white or pink Gaussian noise to the patches in the power domain at a random signal-to-noise ratio (SNR) between 0 and 20 dB. Validation and test patches are excluded from this augmentation. Finally, we normalize the patches. Each patch is z-scored per frequency bin: subtract the per-bin mean, divide by the per-bin standard deviation. The statistics are computed once over all training split frames of one fold.

\subsubsection{Training process}\label{sec214}
We trained one AI system per microphone. For each microphone, the nominal envelope, i.e., "vent"/"vent-low" leak types, "lab"/"work"/"work-low" noise types, no-leak knob state $\leq 3.0$, leak knob state $\geq 5.5$,  is split by whole recording session. This means a leave-one-session-out training with three folds: train on two sessions, validate on the held-out one. Session-level splitting guarantees no data leakage. For microphone-1, the concrete split resulted in 286 training files and 120 validation files, roughly a 70/30 split. The dataset for the training and validation is built per fold. First, the pre-processing is done as mentioned above. Then, the patches get split into 33,580 training patches and 14,080 validation patches for one fold. The normalization statistics are computed on the training split, and applied to both training and validation.

For the training loop, a simple cross-entropy loss is used together with a AdamW \citep{loshchilovDecoupledWeightDecay2017} optimizer. The optimizer has a learning rate of 0.001 and a weight decay of 0.0001. The batch size is 128 and shuffled each epoch. We trained for max 25 epochs with early stopping on the validation loss. The best AI system is saved together with accuracy, macro-F1, confusion matrix, and per-patch probabilities. As it is needed later for the monitoring, the activation taps are extracted for all training and validation patches, i.e., the three post-pool convolutional maps and the 64-d dense embedding. This training loop is repeated for all 3 folds and all 4 microphones resulting in 12 AI systems. The cross-validation results are documented. For all monitoring experiments, fold 3, i.e., training on session 1 and session 2 and validation on session 3, was selected per microphone.

\subsubsection{Hardware}
Training was performed on a workstation with an NVIDIA GeForce RTX GPU with 24GB, using Python 3.12, PyTorch 2.13, and SciPy 1.18. Training a single fold with approximately 34,000 patches and max 25 epochs completed in minutes on the GPU.

\subsection{Coordinated multi-layer runtime monitoring framework}

\begin{figure}
    \centering
    \includegraphics[width=0.8\linewidth]{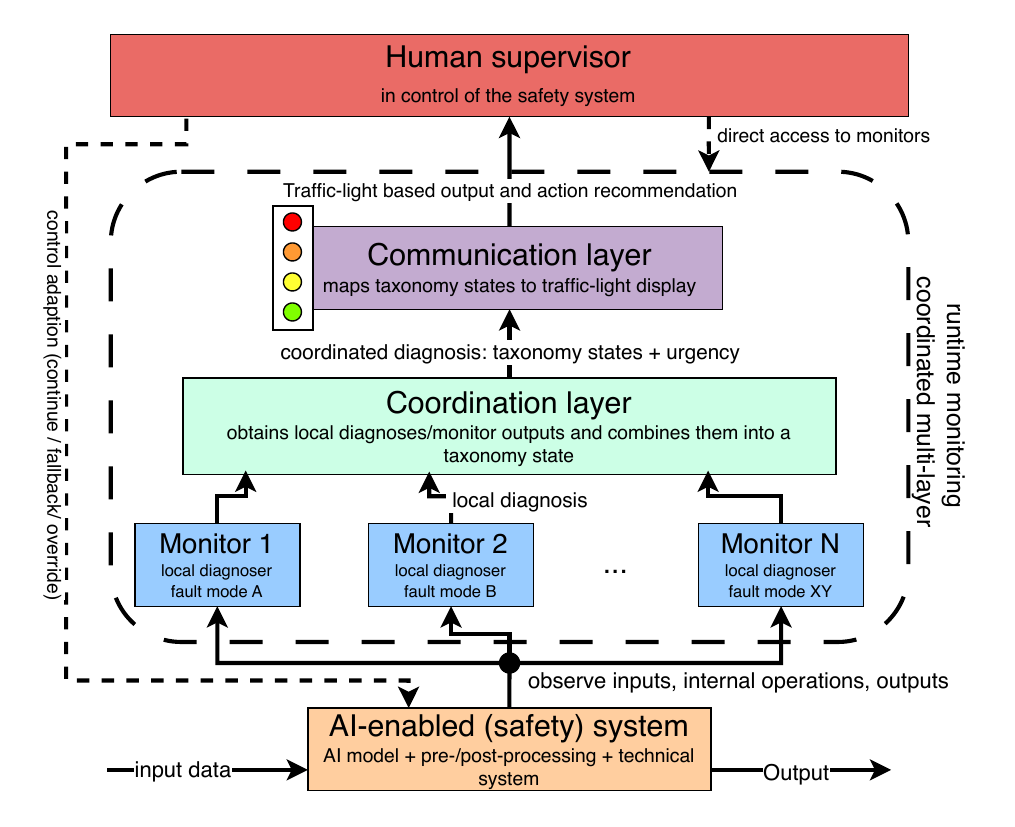}
    \caption{Concept of the COMET framework: monitors act as local diagnosers anchored in distinct fault modes. The coordination layer fuses their local diagnoses into a taxonomy state, which the communication layer presents to the human supervisor as a traffic-light display with action recommendations.}
    \label{fig:comet-framework}
\end{figure}
The framework is derived from two independent sources. The first is pragmatic: single monitors have been shown to produce high false-alarm rates \citep{ferreiraraulsenaRuntimeSafetyMonitoring2023}. The second, stronger source is the theory of fault-tolerant systems. Fault-tolerant control is only possible if faults can be diagnosed and the control adapted accordingly \citep{DiagnosisFaultTolerantControl2006}. In the classical decomposition, fault diagnosis consists of detection, isolation, and identification. In our setting and aligned with \citet{flehmigImplementingArtificialIntelligence2024}, the human supervisor carries the responsibility for control adaptation, and therefore requires diagnosis.

The dominant runtime monitoring technique, OOD detection, provides only detection \citep{yangGeneralizedOutofDistributionDetection2024, luOutofDistributionDetectionTaskOriented2024}. Therefore, we adopt a coordinated diagnosis structure in which each monitor acts as a decentralized local diagnoser to solve a subproblem independently and then the obtained partial diagnoses are combined into a global coordination layer \citep{DiagnosisFaultTolerantControl2006, deboukCoordinatedDecentralizedProtocols2000}. After the coordination layer, a communication layer maps taxonomy states to a traffic-light based display that produces human-centered outputs and promotes quick decision making through simplified taxonomy state representation and recommended actions. Moreover, the human supervisor has the highest control authority and can directly access monitors and their states as well as adapting the control of the AI-enabled safety system by initiating a pre-defined fallback policy or overriding the AI system. The whole concept of the framework is shown in Figure~\ref{fig:comet-framework}.

We use "local" not in a physical meaning but in that sense that a local diagnoser is anchored in and authoritative for a different fault mode. We use the terms "local diagnoser" and "monitor" interchangeably throughout this work. Structurally, the framework is a multi-layer runtime monitor, in line with the layered-protection principle of safety engineering. Functionally, it represents a coordinated diagnosis structure as explained. We retain the term multi-layer runtime monitoring for the framework as a whole, since runtime monitoring is the established term in the AI community, whereas the coordinated-diagnosis reading clarifies the theoretical function.

In this work, we implemented three local diagnosers and they are anchored in signal integrity (L1), physics consistency (L2), and learned competence (L3). The combination extends the diagnosis from detection toward isolation. The coordination layer attributes an event to a fault mode, even though full identification, i.e., magnitude and root cause, remains impractical in complex systems. The coordination and communication layer explained in Section~\ref{sec:comLayer} realizes this coordination by fusing the local diagnoses, and the traffic-light mapping presents the coordinated diagnosis in a form the human supervisor can act on. The following subsections derive the three local diagnosers.

\subsubsection{L1 - Signal-integrity monitor}

The first local diagnoser incorporates five a-priori defined signal-integrity rules. These rules represent faults that are derived from domain knowledge and this is a common technique for fault diagnosis \citep{DiagnosisFaultTolerantControl2006}. Intuitively, the monitor asks whether the audio signal is valid or not. The monitor takes the raw waveform window before it gets transformed with STFT and assumes float audio samples are normalized to full scale $\pm 1$. A binary flag per rule is computed for each window. The following faults are implemented with window $x = (x_1,...,x_N)$ with N = 12,288 samples (256 ms at 48 kHz) and indicator function $1[\cdot]$:
\begin{itemize}
\item 
\textbf{Missing data}: Digital audio is not exactly 0.0 over a longer period, even a quiet room gives noise-floor wiggle. Long stretches of exact zero indicate crashed driver, buffer under-run, a dead network stream. The rule finds the longest continuous zero run and fires it it covers more than 20\% of the window.
\begin{equation}
\text{Fire if} \rightarrow L(x) \geq 0.2 \times N
\end{equation}
With $L(x)$ being the length of the longest contiguous run of exactly-zero samples.
\item 
\textbf{Clipping}: When a signal exceeds the analog-to-digital converter's (ADC) range, the wave-form's top get flattened. This represents that many samples pile up at the same extreme value instead of following a smooth crest. The rule checks whether a suspiciously large fraction of samples sit at $\geq 98\%$ of the window's peak amplitude.
\begin{equation}
\text{Fire if} \rightarrow a \geq 0.2 \land \frac{1}{N} \sum_{i=1}^{N}{1[|x_i| \geq 0.98 \times a]} \geq 0.005
\end{equation}
With peak amplitude $a = max|x_i|$.
\item
\textbf{Flat-line}: A live microphone produces some variance. A constant or near-constant output means the sensor is dead or the ADC is stuck.
\begin{equation}
\text{Fire if} \rightarrow \sigma(x) < 10^{-4}
\end{equation}
With sample mean $\overline{x} = \frac{1}{N}\sum_{i}{x_i}$ and standard deviation $\sigma(x) = \sqrt{\frac{1}{N}\sum_i{x_i - \overline{x}_i}^2}$.
\item
\textbf{DC offset}: Audio is AC-coupled, i.e., sound is pressure variation, so a healthy channel averages to $~0$ over 256 ms. A substantial DC component means a broken bias voltage, a faulty preamp, or a digitization fault.
\begin{equation}
\text{Fire if} \rightarrow |\overline{x}| \geq 0.05.
\end{equation}
\item
\textbf{Level gain anomaly}: A simple check if the overall level is physically plausible. A root mean square (RMS) under $10^{-5}$ is essentially digital silence and a RMS above 0.5 on a $\pm 1$ scale means the channel is driven near its rails that indicates a gain misconfiguration or preamp fault.
\begin{equation}
\text{Fire if} \rightarrow RMS(x) < 10^{-5} \lor RMS(x) > 0.5
\end{equation}
With $RMS(x) = \sqrt{\frac{1}{N}\sum_i{x_{i}^{2}}}$.
\end{itemize}

\subsubsection{L2 - Physics-consistency monitor}

Acoustic leak detection techniques have been studied in almost a century and numerous techniques have been developed \citep{loth2003technology}. These techniques can be compared to consistency-based fault diagnosis where a measurement is or is not consistent with the expect system behavior \citep{DiagnosisFaultTolerantControl2006}. This expect system behavior can be derived from known physics behind a leak. The advantages of using acoustics are fast detection, straightforward retrofitting and easy use. On the other hand, the disadvantage lies in the challenge of high background noise conditions that may mask the actual leak signal \citep{murvay2012survey}. Thus, just using acoustics and their derived physics as fault diagnoser does not generalize well. However, the insights and knowledge we have from the physics are still valuable and should not be ignored when we want to detect leakages. Instead of only using physics and its derived consistency-based fault diagnoser as leak detection, we used the knowledge to create a local diagnoser that checks the physics-consistency between a detected leak from the AI system against known physics.

So, the second local diagnoser incorporates the physics of a compressed-air leakage that produces a characteristic noise. This characteristic noise can be described as broadband, strongly high-frequency-weighted, with a spectral shape that depends on the vent type rather than severity. Intuitively, the second local diagnoser asks whether the detected leak looks physically like a leak. This is done with a fixed spectral template that can be used as evidence to support the AI system's outputs or question them. The physics-consistency diagnoser produces one of four outputs: \textit{consistent}, \textit{quiet}, \textit{inconsistent}, or \textit{abstain}.

\subsubsection*{Offline construction and calibration}
All components of L2, i.e., the spectral leak template, the abstention-gate margin, and the two verdict thresholds, are derived from nominal data only and frozen before deployment. Offline construction and runtime scoring share the same spectral representation. A recording is transformed with the same STFT as the AI system, the power is averaged over time and summed into $b = 8$ log-spaced bands from 100\,Hz to 20\,kHz (100, 194, 376, 729, 1414, 2742, 5317, 10311, 20000\,Hz). This gives us an energy vector $E \in \mathbb{R}^8$ in the linear domain.

The observation model assumes additive contributions, $E_{\mathrm{obs}} = E_{\mathrm{bg}} + E_{\mathrm{leak}}$. For each microphone and observed background condition, the background is estimated as the element-wise median band-energy vector over that condition's no-leak files:

\begin{equation}
E_{\mathrm{bg}}[b] = \operatorname{median}_{j \in \text{no-leak}} E_j[b],
\quad b = 1, \dots, 8
\label{eq:l2-background}
\end{equation}

The median makes the estimate robust to outliers and free of distribution assumptions. In deployment, $E_{\mathrm{bg}}$ would come from a rolling median over recent quiet windows. During evaluation we emulate this with
leave-one-out, i.e., the scored file is excluded from its own background estimate.

The leak template is fitted on the nominal vent-leak training files, separately per microphone and gain configuration. For each vent-leak training file $k$, the excess spectrum against the matching background is computed,
\begin{equation}
r^{(k)}_b = \max\!\left(E^{(k)}_{\mathrm{obs}}[b] - E_{\mathrm{bg}}[b],\; 0\right)
\label{eq:l2-excess}
\end{equation}

and normalized to shares to remove the gain scale:

\begin{equation}
s^{(k)}_b = \frac{r^{(k)}_b}{\sum_{b'} r^{(k)}_{b'}}
\label{eq:l2-shares}
\end{equation}

The template is the element-wise median across files, re-normalized to unit sum:

\begin{equation}
t^{(\mathrm{mic},\mathrm{gain})}_b =
\frac{\operatorname{median}_k s^{(k)}_b}
{\sum_{b'} \operatorname{median}_k s^{(k)}_{b'}},
\qquad \sum_b t^{(\mathrm{mic},\mathrm{gain})}_b = 1
\label{eq:l2-template}
\end{equation}

$t_b$ is the typical fraction of leak excess energy in band $b$. The same vector also represents the abstention-gate weights, $w^{(\mathrm{mic},\mathrm{gain})}_b = t^{(\mathrm{mic},\mathrm{gain})}_b$, since it marks the bands in which leak energy is expected for that configuration. This yields eight independent template/gate pairs (4 microphones $\times$ 2 gains).

The abstention-gate margin $m(\mathrm{mic},\mathrm{gain})$ is calibrated as the 99th percentile of the gate statistic (Eq.~\ref{eq:l2-gate}) on the nominal no-leak files of the corresponding configuration. The two verdict thresholds are calibrated on nominal validation data:

\begin{equation}
\tau_{\mathrm{absent}} = P_{1}(D \mid \text{nominal no-leak}), \qquad
\tau_{\mathrm{present}} = P_{99}(D \mid \text{nominal leak})
\label{eq:l2-thresholds}
\end{equation}

with $D$ as defined in Eq.~\ref{eq:l2-distance}. Only 1\% of no-leak files score below $\tau_{\mathrm{absent}}$, which is the monitor's false-alarm budget. Before freezing, the template's physical validity was verified with three checks on the nominal training data: (i)~\textit{severity scaling}: band energy grows monotonically with knob state, consistent with choked-flow acoustics, (ii)~\textit{shape invariance}: the cosine similarity between per-severity share vectors and the template is $\geq 0.95$ across severity, i.e., severity scales the magnitude but not the spectral shape, (iii)~\textit{separability}: the residual score separates leak from no-leak recordings within each configuration. All checks passed for all configurations. The templates, margins, and thresholds were frozen afterwards and never revisited on held-out data.

\subsubsection*{Runtime}
At runtime, each incoming recording is reduced to its band-energy vector $E^{(i)}_{\mathrm{obs}}$ and the residual against the current background estimate:
\begin{equation}
r^{(i)}_b = \max\!\left(E^{(i)}_{\mathrm{obs}}[b] - E_{\mathrm{bg}}[b],\; 0\right)
\label{eq:l2-residual}
\end{equation}

The abstention gate first checks whether there is enough excess energy in the leak bands to justify a shape comparison at all. The monitor stays mute rather than guessing when the background dominates:

\begin{equation}
\text{abstain if} \quad
10 \ln \frac{\sum_b r^{(i)}_b \, w_b}
{\sum_b E_{\mathrm{bg}}[b] \, w_b} < m(\mathrm{mic},\mathrm{gain})
\label{eq:l2-gate}
\end{equation}

If the gate passes, the residual is normalized to a scale-invariant share vector $s^{(i)}_b = r^{(i)}_b / \sum_{b'} r^{(i)}_{b'}$ and compared against the template by cosine distance:

\begin{equation}
D = 1 - \cos\!\left(s^{(i)},\, t^{(\mathrm{mic},\mathrm{gain})}\right)
= 1 - \frac{s^{(i)} \cdot t}{\lVert s^{(i)} \rVert \, \lVert t \rVert}
\in [0,1]
\label{eq:l2-distance}
\end{equation}

A low $D$ means the excess spectrum has the shape of a vent leak at some severity and a high $D$ means no known leak severity looks similar. The verdict follows from the frozen thresholds:

\begin{itemize}
\item $D < \tau_{\mathrm{absent}}$: \textit{consistent} -- the observation matches
known leak physics; this catches leaks the AI system misses (hidden faults).
\item $D > \tau_{\mathrm{present}}$: \textit{inconsistent} -- no known leak physics
supports the observation; this withholds physical support for the AI system's detection.
\item otherwise: \textit{quiet} -- no strong evidence for either judgment.
\end{itemize}

Intuitively, the two thresholds answer different questions with $\tau_{\mathrm{absent}}$ asking whether there is leak physics the AI system missed, whereas $\tau_{\mathrm{present}}$ asking whether physics contradicts the AI system's detection.

\subsubsection{L3 - Deep k-NN feature-space monitor}

Whereas L2 checks against known physics of leaks, L3 checks against learned competence of the AI system. The L3 local diagnoser is based on using k-nearest neighbors on feature-space embeddings to decide whether a sample is in-distribution (ID) data or out-of-distribution (OOD) data \citep{sunOutofDistributionDetectionDeep2022}. ID data describe the learned data whereas OOD data represent data that are outside the learned competence envelope and potentially lead to unreliable outputs \citep{yangGeneralizedOutofDistributionDetection2024, luOutofDistributionDetectionTaskOriented2024}. In this work, the AI system is based on a CNN model that compresses each patch into a 64-d embedding in the penultimate layer. The idea is if a new patch is embedded far from the data seen during training, then whatever the AI system outputs for it is an extrapolation that may be unreliable and should be flagged. Intuitively, L3 detects and isolates inputs that do not look like something the AI system was trained on.

\subsubsection*{Offline construction and calibration}
First, we constructed and calibrated a threshold that allows us to automatically flag inputs later during runtime. This is done offline before the deployment. All nominal data of the deployed fold are run through the AI system, and the embeddings are extracted for every patch. The embeddings $h \in \mathbb{R}^{64}$ are extracted at the penultimate layer of the CNN that represents a dense output post GELU activation and before the dropout. Next, each embedding is projected onto the unit sphere with an L2-normalization:

\begin{equation}
\hat{h} = \frac{h}{\lVert h \rVert_2}
\label{eq:l2norm}
\end{equation}

This removes the overall activation magnitude and keeps only the direction. Next, the reference store is built by collecting all embeddings of all nominal patches including training, validation, and both classes since nominal leaks are also ID data:
\begin{equation}
R = \{\hat{h}_1, \dots, \hat{h}_N\}, \quad N = 47{,}660 \text{ patches per microphone}
\label{eq:refstore}
\end{equation}

The threshold is calibrated on nominal no-leak recordings $f$ with leave-one-file-out (LOO). All patches of $f$ are removed from the reference store to avoid self-match bias, the k-NN distance score (Eq.~\ref{eq:knn-score}) of each of its patches is computed, and the scores are averaged per recording, $\bar{D}_{\mathrm{kNN}}(f)$. The threshold is the 99th percentile of these file-level scores, targeting a nominal false-alarm rate of 1\%:
\begin{equation}
\theta_{\mathrm{kNN}} = P_{99}\!\left( \bar{D}_{\mathrm{kNN}}(f)
\;\middle|\; f \in \text{nominal no-leak},\ \mathrm{LOO} \right)
\label{eq:knn-threshold}
\end{equation}

Note that the calibration unit is the recording-level mean, which is also the decision unit at runtime.

\subsubsection*{Runtime}

For a new incoming patch, the embedding is extracted and L2-normalized, yielding $\hat{q}$. The Euclidean distance from $\hat{q}$ to all $N$ reference points is computed, and the $k = 50$ smallest are averaged:

\begin{equation}
D_{\mathrm{kNN}}(\hat{q}) = \frac{1}{k} \sum_{r \in \mathcal{N}_k(\hat{q}, R)}
\left\lVert \hat{q} - r \right\rVert_2
\label{eq:knn-score}
\end{equation}

where $\mathcal{N}_k(\hat{q}, R)$ denotes the set of the $k$ nearest reference embeddings. Since all vectors live on the unit sphere, this Euclidean score is a monotone transform of the cosine distance. Using the mean of the 50 nearest neighbors makes the score a local density estimate. The patch scores are averaged per recording to match the calibration unit, and the recording is flagged if

\begin{equation}
\bar{D}_{\mathrm{kNN}}(f) = \frac{1}{n_f} \sum_{i=1}^{n_f}
D_{\mathrm{kNN}}(\hat{q}_i) > \theta_{\mathrm{kNN}}
\label{eq:knn-decision}
\end{equation}

L3 then detects and isolates the input as outside the AI system's competence envelope. It does not say what is wrong or where, since patches are aggregated to keep the false-alarm rate low, but it indicates that the AI system is extrapolating and that its output is potentially unreliable.

\subsubsection{L3 alternative - Deep k-NN-based layer-profile distance monitor}

Whereas L3 compresses the diagnosis into a single number based on the CNN's deepest layer, this alternative local diagnoser keeps the per-layer distances as a 4-d profile and measures how that profile deviates from the profile of normality. The idea builds on the finding that neural networks learn different features at different depths \citep{olahFeatureVisualization2017}. Translated to this work, this means that shifts in the noise (covariate shifts) should affect the early layers, whereas shifts in the decision-relevant concept, such as a novel fault class (semantic shifts), should affect the deeper layers. Monitoring all layers and comparing the resulting profile against the normal profile should therefore not only detect and isolates shifts but also identifies their type.

\subsubsection*{Offline calibration and construction}
As for L3, L2-normalized activations are collected, but not only for the penultimate layer. The activations of each convolutional block after max pooling are tapped as well. The per-layer reference stores are
\begin{equation}
R_l = \{\hat{h}_{l,1}, \dots, \hat{h}_{l,N}\}, \quad
l \in \{\mathrm{conv1}, \mathrm{conv2}, \mathrm{conv3}, \mathrm{dense}\}
\label{eq:refstore-layers}
\end{equation}

built from the same nominal training and validation patches as L3, giving four per-layer reference clouds per microphone. The per-layer k-NN distance is computed with the same leave-one-file-out rule and $k = 50$:
\begin{equation}
D_{\mathrm{kNN}}^{l}(x) = \frac{1}{k}
\sum_{r \in \mathcal{N}_k^{(l)}(\hat{h}_l(x),\, R_l)}
\left\lVert \hat{h}_l(x) - r \right\rVert_2
\label{eq:knn-layer}
\end{equation}

Since raw distances live on incomparable scales across layers (e.g., conv1 distances are systematically smaller than dense distances), they are standardized with the mean $\mu_l$ and standard deviation $\sigma_l$ of the nominal training files' distances:

\begin{equation}
z_l(x) = \frac{D_{\mathrm{kNN}}^{l}(x) - \mu_l}{\sigma_l},
\qquad
z(x) = \left( z_{\mathrm{conv1}}, z_{\mathrm{conv2}}, z_{\mathrm{conv3}},
z_{\mathrm{dense}} \right)
\label{eq:zprofile}
\end{equation}

The 4-d vector $z(x)$ is the z-profile. The prototype of normality is the mean z-profile over all nominal training patches, including both classes:

\begin{equation}
p_z = \operatorname{mean}_{x \in \text{nominal training}} z(x)
\label{eq:prototype}
\end{equation}

Note that $p_z \approx \mathbf{0}$ by construction, since the z-scores are computed with the statistics of the same nominal data. The prototype formalizes the origin of the normalized space rather than adding a separately fitted component.

To avoid noisy per-patch profiles, z-profiles are averaged over blocks of 20 consecutive patches, $\bar{z}_{20}$, and scored by their Euclidean distance to the prototype:

\begin{equation}
D_{\mathrm{profile}} = \left\lVert \bar{z}_{20} - p_z \right\rVert_2
\label{eq:profile-distance}
\end{equation}

The threshold $\theta_{\mathrm{profile}}$ is the 99th percentile of the block scores pooled over the nominal no-leak recordings (LOO), again targeting a 1\% false-alarm rate. The calibration unit is the 20-patch block, matching the runtime decision unit.

\subsubsection*{Runtime}
During the forward pass of the AI system, the four per-layer k-NN distances are computed, standardized to the z-profile, and averaged over 20 consecutive patches. The distance to the prototype (Eq.~\ref{eq:profile-distance}) is compared to the calibrated threshold. If it exceeds $\theta_{\mathrm{profile}}$, a flag is raised. Beyond the flag, the z-profile itself is reported as a diagnostic where covariate shifts are expected to manifest in the early layers, semantic shifts in the dense layer, so the shape of the profile identifies at the shift type.

\subsection{Coordination and communication layer}\label{sec:comLayer}
The design principle is that each local diagnoser of the runtime monitoring framework answers a different question and give answers accordingly:
\begin{itemize}
\item
\textbf{L1}: Is the input signal valid? $\rightarrow$ [OK, fault]
\item
\textbf{L2}: Does the acoustic physics agree with the AI system's detection? $\rightarrow$ [consistent, quiet, inconsistent, abstain]
\item
\textbf{L3}: Is this context inside the competence envelope? $\rightarrow$ [familiar, flag]
\end{itemize}

For L3, two realizations are considered: (i) the dense k-NN distance, and (ii) the layer-profile distance. Both are calibrated to the same 1\% nominal false-alarm rate and frozen before the final evaluation. The decision which of the implementation is used is decided by the evaluation.

The taxonomy is a fixed decision logic that combines the locally obtained diagnoses into one actionable system state. It is the coordination step of the coordinated diagnosis architecture. The local diagnosers are defined during development and frozen before deployment, thus never tuned on the results. The taxonomy has eight different states where each has a specific meaning that is communicated to the human supervisor, see Table~\ref{tab:taxonomy}.

\begin{table}[htbp]
\centering
\caption{Decision taxonomy: system states and their operational meaning.}
\label{tab:taxonomy}
\begin{tabular}{@{}lp{9.5cm}@{}}
\toprule
\textbf{State} & \textbf{Meaning and required action} \\
\midrule
T1-confirmed    & AI system detects leak, physics (L2) consistent $\rightarrow$ trust the detection \\
T1-unconfirmed  & AI system detects leak, physics (L2) abstain $\rightarrow$ report, but mark as unverified \\
T2 (hidden fault) & AI system detects no leak, physics (L2) finds a leak signature (consistent) $\rightarrow$ investigate; the AI system may be blind here \\
T3 (unverified) & AI system detects leak, physics (L2) inconsistent $\rightarrow$ investigate possible novel leak or false alarm; detection lacks physics support \\
T3 (unexplained)  & AI system detects no leak, physics (L2) inconsistent $\rightarrow$ investigate; unexplained physical anomaly \\
T4 (deferred)   & Input outside the competence envelope (L3) $\rightarrow$ the AI system is extrapolating; defer to the operator \\
all-clear       & AI system detects no leak, physics (L2) quiet or abstained $\rightarrow$ normal operation \\
L1 override     & Invalid input signal (L1) $\rightarrow$ the pipeline abstains entirely; sensor/digitization fault \\
\bottomrule
\end{tabular}
\end{table}

The order of the multi-layer runtime monitoring matters. A L1 override blocks the downstream output because no learned or physics judgment is meaningful on a corrupted signal. A T2 represents a physics-positive finding on an AI system misclassification and combined with a L3 flag that is the most safety-critical where the AI system misses a physics-consistent leak and is outside its competence envelope. Translated into a traffic-light representation for human supervisors based on the extended framework of ISO/IEC TR 5469 \citep{flehmigImplementingArtificialIntelligence2024}, this would look like in Table~\ref{tab:trafficlight}.

\begin{table}[htbp]
\centering
\caption{Traffic-light operator display: urgency mapping of the taxonomy states.}
\label{tab:trafficlight}
\begin{tabular}{@{}lp{10cm}@{}}
\toprule
\textbf{Level} & \textbf{States and required action} \\
\midrule
Green  & T1-confirmed, all-clear $\rightarrow$ informed trust \\
Yellow & T1-unconfirmed, L1 override $\rightarrow$ caution, verification needed \\
Orange & T2, T3 (unverified), T3 (unexplained), T4 $\rightarrow$ operator decision required \\
Red    & T2 with concurrent L3 flag $\rightarrow$ immediate attention, fallback policy \\
\bottomrule
\end{tabular}
\end{table}

The states are compressed into a four-level traffic light, with red reserved for the compound case of a hidden fault inside an unfamiliar context. The aim is to improve the comprehensibility of the coordinated multi-layer runtime monitoring while still enabling the human supervisor to dig deeper into flags if needed.

\section{Experiments}
The central premise of this experimental evaluation is that for safety-critical deployments, distribution shifts are partly known unknowns, i.e., we must expect that covariate shifts such as changing background noise, and semantic shifts such as new leak types will occur during runtime but we do not know when and how they will look like. Any training set, however large, cannot cover all future operating conditions. A monitoring system must therefore be trained and calibrated entirely on the data available at deployment time, and prove its worth by detecting deviations it was never shown before.

Our experiment is designed to reproduce this situation. We simulate the deployment setting by restricting all training, calibration, and design decisions to a nominal envelope, while deliberately holding out conditions that stand in for the different shift types. The chosen dataset is suited for this with its hydraulic-noise background provide a covariate shift, and its tube leaks provide a semantic shift. Because these conditions were never touched during development, the experiment emulates the real-world asymmetry between what is known at deployment and what must be detected at runtime, including its hardest case where shift whose exact form is not known in advance. A third held-out condition, synthetically injected sensor faults, probes the complementary case of shifts that are almost fully specifiable a-priori and therefore addressable by fixed integrity rules.

\subsection{Held-out shift conditions}
We evaluate on the IDMT-ISA-Compressed-Air dataset \citep{johnsonIDMTISACompressedAirDataset2020} with a CNN-based AI system and the training as described in Section~\ref{sec21}. There are three shift types in the data that were never touched during development:

\begin{itemize}
\item
\textbf{Covariate shift}: recordings on the hydraulic-noise backgrounds where the leak classes are known but the acoustic environment is not.
\item
\textbf{Semantic shift}: tube-leak recordings with all severity and all nominal backgrounds where a novel fault class occurs within a familiar environment.
\item
\textbf{Sensor faults}: synthetic waveform faults injected into nominal patches at evaluation time with parameters fixed a-priori such as dropout (continuous 0 over 50\% of patch), clipping (hard clip at $\pm 50\%$ full scale), flat-line (channel replaced by its mean), and DC offset ($\pm0.2$).
\end{itemize}

As a sanity anchor, nominal session-3 validation files are scored alongside. A deviation of their flag rates from the calibrated values is treated as a bug report, not a result.

\subsection{Freeze settings after development}
All monitor design, calibration, and threshold decisions were made exclusively on nominal data and then frozen. The held-out conditions were evaluated exactly once against this frozen specification. Pre-registered sensitivity variants such as k-NN with $k = \{10,50,100\}$, class-conditional Mahalanobis distance \citep{mahalanobis2018generalized} and maximum softmax probabilities (MSP) \citep{hendrycksBaselineDetectingMisclassified2016}, are reported descriptively alongside the confirmatory main specification. Components that failed their pre-registered acceptance criteria during development like supervised-contrastive retraining to improve the deep k-NN feature-space monitor, and windowed alarm aggregation are reported as negative results.

\subsection{Evaluation metrics}\label{sec33}
Per microphone and condition we report the accuracy and recall of the AI system and the false-alarm rate, per-monitor outputs, i.e., L1 rule rates, L2 verdict distribution, L3 k-NN and layer-profile flag rates and mean per-layer z-profiles, the taxonomy-state and traffic-light distributions of the coordinated multi-layer runtime monitoring (COMET). Human-supervisor-oriented metrics are evaluated through an ablation ladder of monitoring configurations:
\begin{itemize}
\item
\textbf{S0}: bare AI system.
\item
\textbf{S1a/b/c}: AI system plus a single monitor, i.e., MSP-only, L2-only, L3-only.
\item
\textbf{S2}: COMET.
\end{itemize}

The reporting includes the trusted-wrong-answer rate, i.e., decisions presented as reliable that are wrong, missed-leak coverages, i.e., system state conditioned on true leaks the AI system did not detect, including confident misses at max-softmax $\geq 0.9$, and the nominal cost, i.e., false-alarm rate of each configuration on nominal no-leak.

\section{Results}
\subsection{Baseline performance and AI system behavior under shift}
Three disjoint data roles are used throughout this section.
\begin{itemize}
\item
\textbf{Training data}: nominal sessions 1–2, used to train the AI system. This data includes both leak types "vent" and "vent-low" on backgrounds "lab", "work", and "work-low".
\item
\textbf{Nominal validation data}: the nominal session-3 recordings, used during development for model selection and for calibrating every monitor thresholds. Its patch embeddings also form part of the k-NN reference store and the profile reference store.
\item
\textbf{Held-out evaluation data}: the hydraulic backgrounds "hydr"/"hydr-low", tube leaks, and injected sensor faults, which were never touched until the single post-freeze evaluation.
\end{itemize}

The deployed per-microphone AI systems achieve patch-level accuracies of $94.2-96.7\%$ measured on the nominal validation data. The macro-F1 scores $94.2-96.4\%$ and cross-validates over all three folds the mean accuracy per microphone is $95.3-97.5\%$. At file level, the AI systems produce zero false alarms and a leak recall of $93.1\%$. These numbers are the reference point for the shift results below. We note that this baseline required noise augmentation, AI systems trained without it scored $>95\%$ under clean lab conditions as documented in \citet{johnson2020compressed}.

On the nominal validation data, the frozen pipeline reproduces its calibration values. The dense k-NN flags $0.0\%$ of no-leak recordings, and the layer-profile monitor flags $1.9\%$. The only deviation is an L1 defect where the clipping rule over-fires on microphone-1 $15\%$ of its nominal no-leak files, versus $0-2.5\%$ on the other microphones. This lifts the pooled L1 rate on nominal no-leak files to $3.7\%$ which will be discussed later in this section.

Next, we ran the AI system under shift. Without monitoring, the classifier's behavior on the held-out conditions demonstrates the problem this work addresses. Under covariate shift, the AI system does not detect any leak on the low-intensity hydraulic background "hydr-low", i.e., $0\%$ false alarms on no-leak files, but produces false alarms on $37.5\%$ of no-leak files on the loud hydraulic background "hydr". These false alarms are confident, wrong, and invisible from the output alone. Under semantic shift, recall drops from $93.1\%$ to $69.8\%$ on tube leaks, and this number overstates the AI system's competence, since the tube leak class was never part of training. The correct-looking verdict are extrapolations, and the $30.2\%$ misses are silent. On faulted audio, the AI system still detects leaks on $25-69\%$ of files, so decisions computed on faulty input. In all three cases, a human supervisor reading the bare AI system output would be systematically misinformed.

\subsection{Per-monitor behavior on the held-out shifts}
\subsubsection{L1 - Signal-integrity monitor on injected faults}\label{sec421}
L1 detects and isolates $100\%$ of dropout, flat-line, and offset injections and only $9.6\%$ of clipping injections. Every L1-positive file was routed to the override state, meaning the AI system and downstream monitors were blocked on exactly the files where their outputs would have been unreliable. Without this override, the AI system would have detected leaks on $25-69\%$ of faulted files. Notably, the k-NN monitor (L3) would independently have flagged $98\%$ of dropout and $100\%$ of flat-line files, so the integrity layer's veto is even backed by the feature-space layer but not for clipping, where L1's blind spot propagates.

\subsubsection{L2 - Physics-consistency monitor under covariate and semantic shift}
\begin{table}[htbp]
\centering
\caption{L2 physics-layer verdict distribution (\%) on nominal leak validation data, per microphone and pooled. "Blocked" denotes files on which an L1 integrity rule fired. On non-injected conditions, blocking is dominated by the microphone-1 clipping-rule defect.}
\label{tab:l2nominalleak}
\begin{tabular}{@{}lccccc@{}}
\toprule
\textbf{verdict} & \textbf{mic1} & \textbf{mic2} & \textbf{mic3} & \textbf{mic4} & \textbf{pooled} \\
\midrule
consistent    & 57.5\,\% & 33.8\,\% & 62.5\,\% & 35.0\,\% & 47.2\,\% \\
quiet         & 5.0\,\%  & 0\,\%    & 16.3\,\% & 0\,\%    & 5.3\,\% \\
inconsistent   & 0\,\%    & 0\,\%    & 0\,\%    & 0\,\%    & 0\,\% \\
abstain       & 26.3\,\% & 66.2\,\% & 21.2\,\% & 65.0\,\% & 44.7\,\% \\
blocked (L1)  & 11.3\,\% & 0\,\%    & 0\,\%    & 0\,\%    & 2.8\,\% \\
\bottomrule
\end{tabular}
\end{table}

The physics-consistency monitor's defining result is its calmness under covariate shift. On hydraulic no-leak files it issues zero \textit{consistent} verdicts with 0\% physics-supported false alarms, precisely where the AI system produces its confident errors. Its dominant verdict on the loud shifted backgrounds is \textit{abstain} with 83--88\%. The template-weighted excess gate (Eq.~\ref{eq:l2-gate}) mutes the shape comparison where the background dominates the leak bands, so the monitor degrades to silence rather than to false reassurance.

On nominal leaks (Table~3), L2 confirms 47.2\% of files and abstains on 44.7\%, with zero \textit{inconsistent} verdicts. It supports correct detections where the signal is strong enough and never withholds support from a true leak. On tube leaks (Table~4), it abstains on 65\% and judges only 1.4\% explicitly \textit{inconsistent}. The inconsistency almost never fires, so semantic-shift detection is carried by the feature-space layer (Section~4.2.3). One anomaly stands out where microphone-3 issues 29.9\% \textit{consistent} verdicts on tube leaks, while the other microphones issue essentially none. This deviating template fit produces the only trusted tube confirmations and is examined in Section~4.6.3.

\begin{table}[htbp]
\centering
\caption{L2 physics-layer verdict distribution (\%) on tube-leak files, pooled and per microphone.}
\label{tab:l2verdicts}
\begin{tabular}{@{}lccccc@{}}
\toprule
\textbf{L2 verdict} & \textbf{pooled} & \textbf{mic1} & \textbf{mic2} & \textbf{mic3} & \textbf{mic4} \\
\midrule
abstain       & 64.8\,\% & 50.7\,\% & 70.8\,\% & 63.9\,\% & 73.6\,\% \\
quiet         & 22.6\,\% & 35.4\,\% & 28.5\,\% & 0.7\,\%  & 25.7\,\% \\
consistent    & 7.5\,\%  & 0\,\%    & 0\,\%    & 29.9\,\% & 0\,\% \\
inconsistent   & 1.4\,\%  & 0\,\%    & 0\,\%    & 5.6\,\%  & 0\,\% \\
blocked (L1)  & 3.8\,\%  & 13.9\,\% & 0.7\,\%  & 0\,\%    & 0.7\,\% \\
\bottomrule
\end{tabular}
\end{table}

\subsubsection{L3 - Feature-space monitor k-NN vs layer-profile}

\begin{figure}
    \centering
    \includegraphics[width=0.6\textwidth]{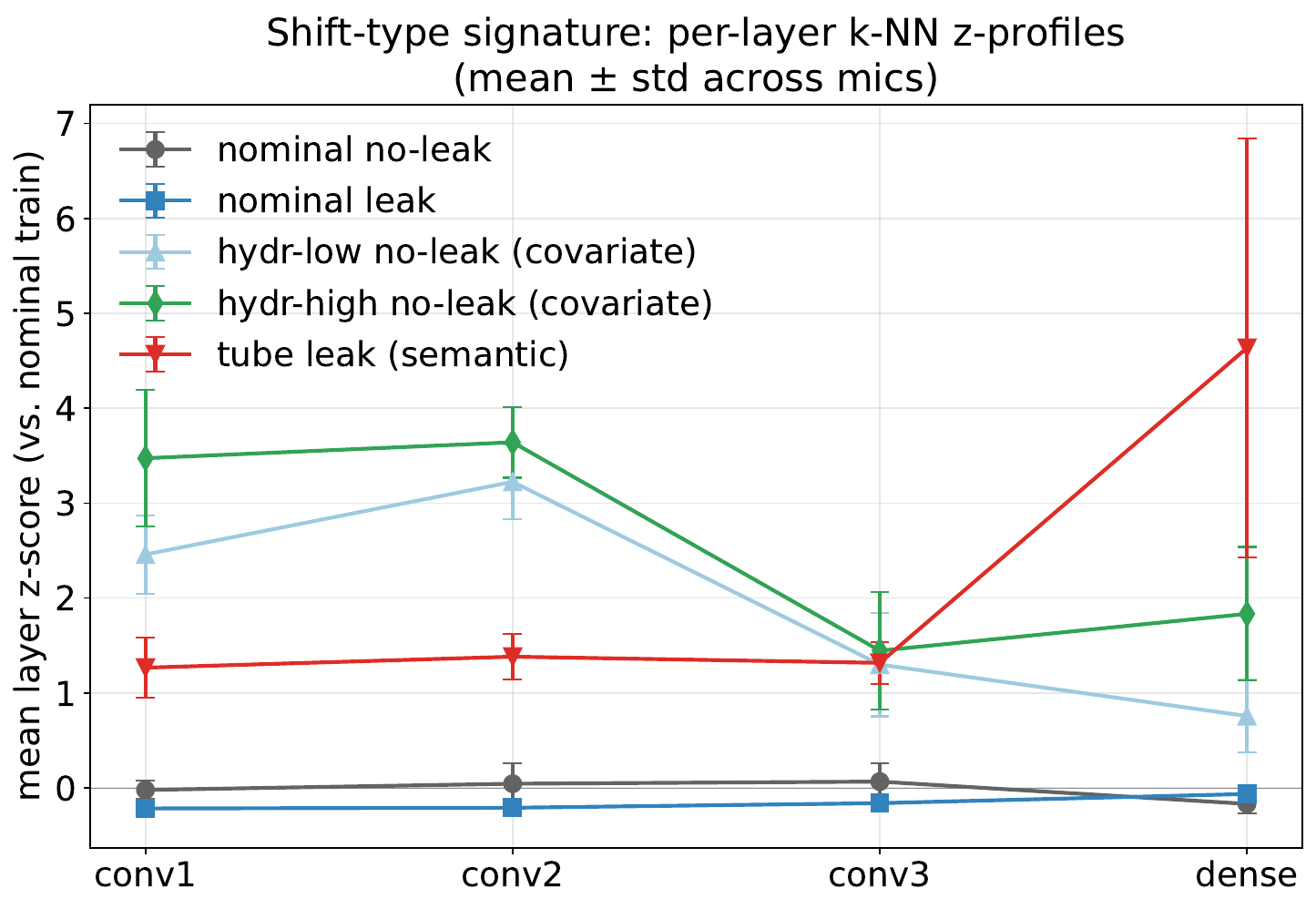}
    \caption{Layer-wise shift signatures. Mean z-standardized k-NN distance profile (conv1–dense) per condition, averaged over microphones. The covariate shift (hydraulic backgrounds) peaks at the early convolutional layers, the semantic shift (tube leaks) at the dense layer, and the nominal control remains at $\approx 0$ throughout. This confirms that shift type localizes by network depth, and explaining the complementary sensitivities of the dense k-NN and the layer-profile monitor.}
    \label{fig:zprof}
\end{figure}

Both realizations behave as their calibration predicts and as the layer-localization hypothesis expects. The dense k-NN flags $0\%$ of nominal no-leak files, $32-54\%$ of hydraulic no-leak files (covariate), and $74.5\%$ of tube leaks (semantic). The layer-profile monitor is more sensitive to the covariate shift with $89.2\%$ on "hydr-low" and $98.6\%$ on "hydr" no-leak files at a slightly higher number of false alarms $1.9\%$. It almost matches the k-NN on the tube leaks with $69.3\%$. We hypothesized that shift type localizes by network depth, i.e., covariate shifts, which alter low-level input statistics, manifest in early convolutional feature distances, while semantic shifts, which present concepts absent from training, manifest in the task-oriented deep layers. The per-layer z-profiles, see Figure~\ref{fig:zprof}, confirm this pattern the hydraulic shift peaks at conv1/conv2, the tube shift spikes at the dense layer (semantic), nominal sits at $\approx0$. The profile's sensitivity comes at a cost with $1.9\%$ false positives on nominal no-leak files and, $34.4\%$ of nominal leak files compared to $0.0\%$ and $5.6\%$ from the k-NN. These files represented the trained in-distribution data. The high number of false alarms on correct leak detection on ID data makes it difficult to implement the layer-profile as main L3 monitor. Therefore, we use the dense k-NN as the L3 vote and use the layer-profile monitor as diagnostic complement. Its z-profile signature and its covariate sensitivity inform the analysis without entering the fusion rule.

\subsection{Taxonomy states and traffic lights under shift}
On the nominal operation, COMET and its communication layer is quiet and green. $95.6\%$ of no-leak files reach all-clear where the remainder is the microphone-1 clipping artifact. On nominal leaks, $81.6\%$ are presented as trusted detections, with $46.6\%$ T1-confirmed where the AI system detects a leak and physics (L2) agrees and $35.0\%$ T1-unconfirmed where the AI system detects a leak and physics are abstained. The remaining leaks are deferred (T4) with $5.6\%$ and $4.7\%$ are missed silently. The trusted-wrong-answer rate on nominal data is $0\%$.

\subsubsection{Covariate shift}
COMET separates "novel context" from "novel fault" exactly as designed. On "hydr-low" no-leak files, where the AI system itself stays silent, $67.7\%$ reach all-clear and $19.8\%$ are flagged T4 as unfamiliar context, with zero false leak alarms. On "hydr" background, where the bare AI system false-alarms on $37.5\%$ of no-leak files, COMET converts these into safe states. The L1 overrides $46.5\%$ as the gain rule treats the extreme background level as an integrity anomaly and $17.4\%$ T4 with zero trusted false alarms. On hydraulic leak files, detections are predominantly presented as T1-unconfirmed or T4 rather than confirmed, i.e., COMET reports the leak but withholds its guarantee in the unfamiliar context. We note that had the layer-profile flag driven the T4 state instead of the dense k-NN, the deferred share on hydraulic no-leak files would rise from $19.8\%$ to $\approx 89\%$. However on the nominal data, the same change would divert $34.4\%$ of leak files from trusted states.

\subsubsection{Semantic shift}
The dominant state is T4 with $70.1\%$ and an orange light. The k-NN recognizes the unfamiliar context and defers the verdict. This reflects the desired behavior where a never-seen leak type is deferred to the human supervisor, and it covers the majority of files regardless of whether the AI system happened to vote leak. However, $8.3\%$ of the tube-leak files remain all-clear, i.e., no leak detection, physics abstaining, and feature space unfamiliar only mildly. Trusted confirmations occur only via the microphone-3 template anomaly with $2.3\%$.

\subsubsection{Sensor faults}
Dropout, flat-line, and offset injections are intercepted entirely bu the signal-integrity layer (L1). L1 overrides $100\%$ of these files and blocks downstream layers, avoiding the $25-69\%$ of AI system verdicts on faulty inputs. Clipping is the exception that marks the system's boundary where only $9.6\%$ are caught, and $63.5\%$ of clipped files proceed to a trusted state. This is a direct, measurable consequence of the L1 blind spot reported in Section~\ref{sec421}.

\subsubsection{Traffic-light view}
\begin{figure}
    \centering
    \includegraphics[width=0.8\textwidth]{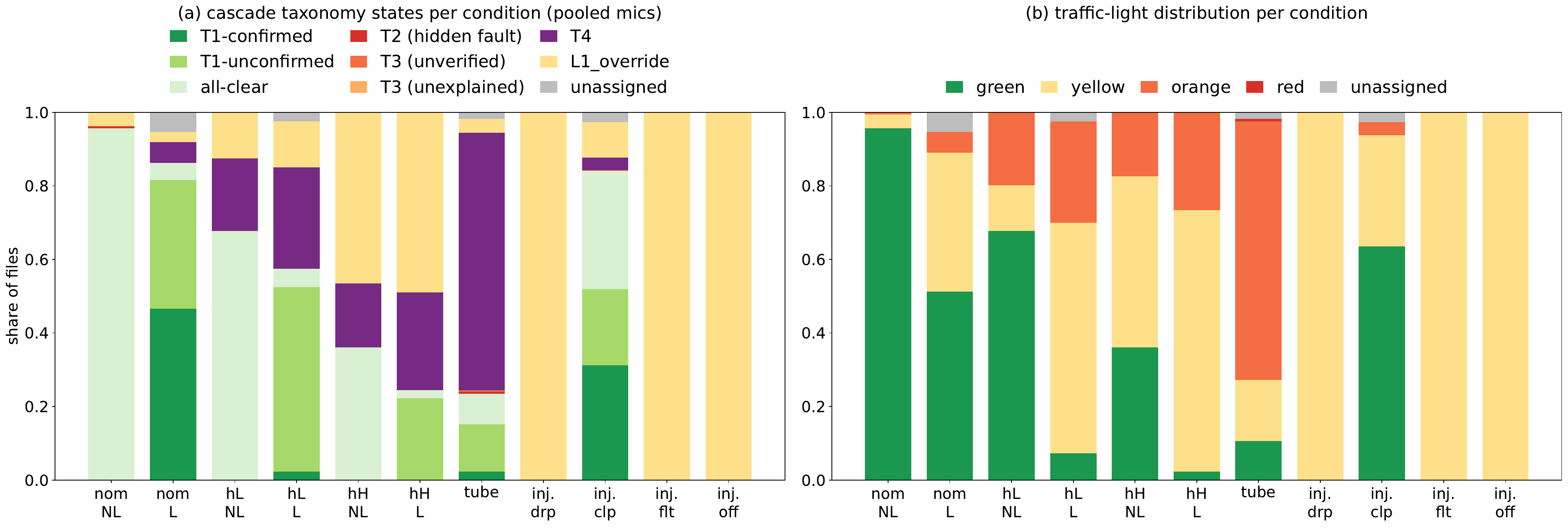}
    \caption{Cascade verdicts per condition pooled over microphones. (a) Distribution of taxonomy states: nominal no-leak data is dominated by all-clear and nominal leaks by trusted detections (T1-confirmed/unconfirmed). The covariate shift (hydraulic backgrounds, hL/hH) splits between all-clear, integrity override, and deferred verdicts (T4). The semantic shift (tube leak) is dominated by T4. Injected faults (inj.) are intercepted by the L1 override (dropout, flat-line, offset) except clipping, which exposes the integrity layer's blind spot. (b) The corresponding traffic-light distribution a human supervisor would see: nominal operation is overwhelmingly green, shifts escalate to yellow/orange, and sensor faults are held at yellow. No wrong condition is presented as trusted green, except for the clipping blind spot and the residual tube silent misses.}
    \label{fig:trafficView}
\end{figure}
As seen in Figure~\ref{fig:trafficView}, pooled over microphones, the signal a human supervisor would see contains nominal no-leak $95.6\%$ green, "hydr"/"hydr-low" no-leak $67.7\%$ and $36.1\%$ green with the rest escalated to yellow and orange rather than falsely trusted, tube leaks are $70.3\%$ orange, and sensor faults like dropout, flat-line, and offset are $100\%$ yellow. No condition produces red, i.e., hidden faults and unfamiliar context, beyond isolated files on tube leaks with $0.7\%$.

\subsection{Ablation ladder}
\begin{figure}
    \centering
    \includegraphics[width=0.8\textwidth]{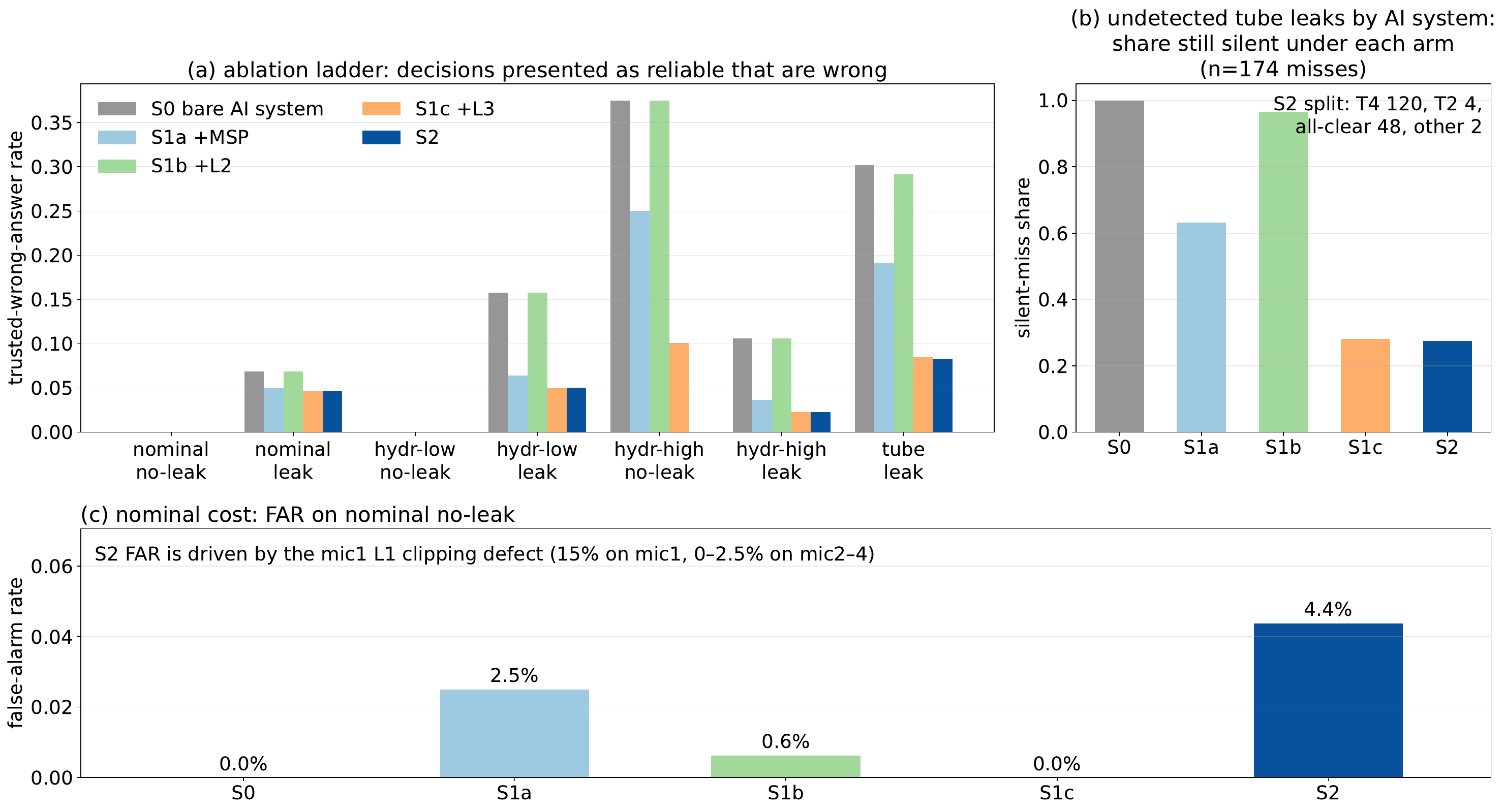}
    \caption{Monitoring ablation ladder pooled over microphones. (a) Trusted-wrong-answer rate per condition for the bare AI system (S0), single-monitor arms (S1a: +MSP, S1b: +L2, S1c: +L3), and COMET (S2). (b) Missed-leak coverage on tube leaks the AI system called negative (n =174). (c) Nominal cost: false-alarm rate of each arm on nominal no-leak data. COMET's $4.4\%$ is driven by the microphone-1 clipping-rule defect.}
    \label{fig:TWA}
\end{figure}

We compare five systems per condition as presented in Section~\ref{sec33}. The trusted-wrong-answer (TWA) rate, missed-leak coverage, and nominal false-alarm cost are used and displayed in Figure~\ref{fig:TWA}. TWA rate describes the share of files presented as reliable whose verdict is wrong.

S0 trusts every output by construction, thus TWA reaches $37.5\%$ on "hydr" no-leak, $30.2\%$ on tube leaks, and $58.3\%$ on flat-line faults. $12-76\%$ of these errors are confident with max-softmax $\geq 0.9$. S1a reduces TWA only moderately (tube $30.2 \rightarrow 19.1\%$, flat-line $58.3 \rightarrow 41.7\%$). It also misses most faulted-audio input with a flag rate of $1.5-25\%$ on injected-fault files, on which the AI system still detects leaks $25-69\%$ of the time, thus confidence is not a sufficient detector of invalid input. S1b barely moves TWA (tube $30.2 \rightarrow 29.2\%$). It is mostly abstained and withholds its judgment exactly where the AI system fails, thus $96.6\%$ of AI system misclassifications on true leaks the monitor stays silent. S1c is the strongest single monitor (tube $30.2 \rightarrow 8.5\%$, flat-line $0\%$) but cannot distinguish what is wrong, and on "hydr" no-leak it still leaves $10.1\%$ trusted-wrong.
S2, COMET, reduces TWA to $0\%$ on "hydr" no-leak, $2.3-5.0\%$ on hydraulic leak conditions, $8.3\%$ on tube, and $0\%$ on dropout/flat-line/offset. In each case equal to or better than the single monitor, because the layers cover each other's blind spots. COMET's advantage is not only quantitative, its coordinated multiple monitor structure turns a bare detection flag into a diagnosis with detection and isolation, whereas a single monitor such as S1c can only detect that something is off. L1 intercepts sensor faults, L2 withholds physical support or confirms physics, L3 catches unfamiliar context. The one condition where S2 is not clean is tube leaks. Of the 174 true-leak files the AI system misclassified, 120 are caught as T4 and 4 as T2, but 48 ($27.6\%$) remain all-clear silent misses. This is the system's worst case. On clipping faults, S2's residual TWA of $3.1\%$ is inherited directly from L1's blind spot.

The cost of human supervision with a coordinated multi-layer runtime monitoring such as COMET is small and known. On nominal no-leak data the false-alarm rate is $0\%$ for S0 and S1c, $0.6\%$ for S1b, $2.5\%$ for S1a, and $4.4\%$ for S2. The latter is driven entirely  by the microphone-1 clipping-rule defect. Without that defect COMET's nominal cost would be $\approx1\%$.

\begin{table}[t]
     \centering
     \caption{Ablation ladder: trusted-wrong-answer (TWA) rate (\%) per condition and system arm, pooled over the four microphones. S0: bare AI system; S1a: +MSP confidence; S1b: +L2 physics; S1c: +L3 k-NN; S2: COMET. Bottom rows: silent-miss share on tube leaks the AI system did not detect ($n=174$) and nominal false-alarm cost (nominal no-leak). Bold: best per row.}
     \label{tab:ablation}
     \begin{tabular}{ll rrrrr}
     \toprule
     & & S0 & S1a & S1b & S1c & S2 \\
     & & bare & +MSP & +L2 & +L3 & COMET \\
     \midrule
     \multicolumn{7}{l}{\emph{Nominal data}}\\
     & no-leak & 0.0 & 0.0 & 0.0 & 0.0 & \textbf{0.0} \\
     & leak    & 6.9 & 5.0 & 6.9 & 4.7 & \textbf{4.7} \\
     \midrule
     \multicolumn{7}{l}{\emph{Covariate shift (hydraulic background)}}\\
     & hydr-low no-leak  & 0.0  & 0.0 & 0.0  & 0.0 & \textbf{0.0} \\
     & hydr-low leak     & 15.8 & 6.4 & 15.8 & 5.0 & \textbf{5.0} \\
     & hydr no-leak & 37.5 & 25.0 & 37.5 & 10.1 & \textbf{0.0} \\
     & hydr leak    & 10.6 & 3.6 & 10.6 & 2.3 & \textbf{2.3} \\
     \midrule
     \multicolumn{7}{l}{\emph{Semantic shift (tube leak)}}\\
     & tube leak & 30.2 & 19.1 & 29.2 & 8.5 & \textbf{8.3} \\
     \midrule
     \multicolumn{7}{l}{\emph{Injected sensor faults}}\\
     & dropout  & 14.8 & 13.1 & 13.8 & 1.0 & \textbf{0.0} \\
     & clipping & 4.8  & 3.3  & 4.8  & \textbf{3.1} & \textbf{3.1} \\
     & flat-line & 58.3 & 41.7 & 58.3 & \textbf{0.0} & \textbf{0.0} \\
     & offset   & 4.6  & 3.7  & 4.6  & 0.8 & \textbf{0.0} \\
     \midrule
     \multicolumn{2}{l}{Tube silent misses (\% of 174)}
               & 100.0 & 63.2 & 96.6 & 28.2 & \textbf{27.6} \\
     \multicolumn{2}{l}{Nominal false-alarm cost (\%)}
               & \textbf{0.0} & 2.5 & 0.6 & \textbf{0.0} & 4.4$^{\dagger}$ \\
     \bottomrule
     \multicolumn{7}{l}{\footnotesize $^{\dagger}$driven by the mic1 clipping-rule
     defect (15\,\% on mic1; 0--2.5\,\% on mic2--4).}
     \end{tabular}
     \end{table}

\subsection{Sensitivity variants}
Alongside the main specification, we report the pre-registered variants, i.e. different neighborhood size $k$, Mahalanobis distance, and output-confidence MSP, at their thresholds. For $k=10$, the L3 monitor gets slightly more sensitive with $79.3\%$ on tube leaks, and $65.6\%$ on "hydr" no-leak files. On the other hand, $k=100$ is more conservative with $69.6\%$ and $51.0\%$. The gap between shift types are unchanged, so the monitor's behavior does not hinge on k. The class-conditional Mahalanobis distance is consistently less sensitive than the k-NN (L3) with $55.7\%$ vs $74.5\%$ on tube leaks and $41.3\%$ vs $53.8\%$ on "hydr" no-leak files and less clean in terms of false alarms ($3.8\%$ vs $0\%$ on nominal no-leak files). MSP is a weak shift detector with only $15.6\%$ on tube leaks, barely above its nominal-leak rate of $2.8\%$. This confirms that a AI system's confidence in its answer is a poor proxy for the familiarity of its input, which is precisely the gap the feature-space monitor fills.

\subsection{Negative results and limitations}
\subsubsection{Rejected development branches}
Two pre-registered approaches failed their acceptance criteria and were discarded before the evaluation. (i) Supervised-contrastive retraining of the encoder, intended to enrich the penultimate embedding \citep{sunOutofDistributionDetectionDeep2022}, but did not raise the participation ratio and was rejected. (ii) Windowed alarm aggregation, i.e., alarm if $\geq 25\%$ of windows in a sliding 20-windows block flag. This produced nominal false-alarm rates of $4-19\%$ so the recording-level mean was kept, with the consequence that the calibrated decision unit is the $~30$ s recording.

\subsubsection{Integrity-layer boundaries}
L1's clipping rule has a measured blind spot with soft clipping at $\pm 50\%$ of full scale is detected on only $9.6\%$ of files, and $63.5\%$ of clipped files consequently proceed to a trusted state. The same rule over-fires on microphone-1, inflating COMET's nominal cost to $4.4\%$. Both defects are specification-level, not learned and thus fixable by rule revision.

\subsubsection{Physics-consistency monitor conservatism}
L2's inconsistency threshold is deliberately conservative, and on tube leaks the inconsistency almost never fires. The semantic shift detection is carried by the feature-space layer, not by physics. The per-microphone template fit also matters where microphone-3's deviating template is responsible for all spurious "consistent" verdicts on tube leaks and the only trusted tube confirmations.

\subsubsection{Residual worst case}
48 of 576 tube-leak files remain all-clear, i.e., AI system does not detect a leak, physics abstaining, feature space close enough to the reference, including 22 files the AI system does not detect a leak with high confidence. COMET reduces, but does not eliminate, silent misses under semantic shift.

\section{Discussion}
The central finding of this work is that no single monitor or local diagnoser was sufficient, and that the fault modes of the individual monitors were largely disjoint as it was intended. Confidence-based monitoring reduced but did not remove trusted wrong answers. The physics-consistency monitor withheld its judgment exactly where it was most needed, and the feature-space monitor was the strongest single monitor but could not isolate and identify the fault. Only COMET reduced the trusted-wrong-answer rate to zero on hydraulic no-leak data and on three of four injected sensor faults, at a nominal false-alarm rate of $\approx 1\%$. Equally important the qualitative difference that the coordinated multi-layer runtime monitoring structure with coordination and communication layers turns a detection flag into a diagnosis with detection, isolation and identification for some fault modes, i.e., sensor fault, physically implausible detection, or unfamiliar operating context. This is the form of information a responsible human supervisor can act on and adapt to.

\subsection{Need for runtime monitoring}
The results confirm that runtime monitoring of AI systems is necessary to catch silent faults. The bare AI system produced confident, wrong, and invisible outputs on every shifted condition. 37.5\% false alarms under loud hydraulic noise, 30.2\% silent misses on tube leaks, and leak verdicts on 25--69\% of files with corrupted input. These errors cannot be detected by output-level inspection, since 12--76\% of them were confident (max-softmax $\geq 0.9$). That monitoring is needed is not new. It is already standard practice in MLOps \citep{kreuzbergerMachineLearningOperations2023}. However, MLOps monitoring works through periodic assessments with slow feedback loops, whereas a safety system must act on potentially unreliable output immediately, at decision time. Monitors for this setting must therefore be not only reliable but also informative for the human supervisor, and, as \citet{ferreiraraulsenaRuntimeSafetyMonitoring2023} showed, well calibrated, since high false-alarm rates would be counterproductive.


\subsection{Experimental design}
The held-out conditions stand in for a real deployment challenge where training data are known to be incomplete, but it is not known how the world will change, \textit{known unknowns}, as \citet{hendrycksIntroductionAISafety2024} calls them. Our experimental design aims to mirror this asymmetry. All training, calibration, and design decisions used nominal data only, and the shifts were evaluated once against the frozen specification, a zero-shot test of the monitors. One point is open to discussion that is the categorization of the held-out conditions into covariate and semantic shift is our assignment, made a-priori from domain knowledge. However, the measured z-profile signatures support it empirically. The hydraulic shift peaks in the early convolutional layers, where low-level input statistics are represented, while the tube-leak shift peaks in the dense, task-oriented layer and the AI system's degraded performance confirms that both conditions were genuine distribution shifts for the model. 

What the experiment cannot cover are \textit{unknown unknowns}, which may appear during runtime but are impossible to model in advance. The COMET's concept of providing multiple layers of monitoring that are anchored on different fault modes is intended to be robust to such cases. In this work the proposed layers were anchored on validity, physics, and familiarity. Overall, we consider the experiment a proof-of-concept under realistic asymmetry between what is known at deployment and what must be detected at runtime.

\subsection{Human supervisor}
Our overall system design follows \citet{flehmigImplementingArtificialIntelligence2024} and \citet{flehmigPerspectivesReliabilityMonitoring2026}, where the human supervisor remains in control of the AI-enabled safety system and decide whether to trust its output, switch to a fallback policy, or overrides it. To adequately perceive, interpret, and decide in such an event, the supervisor needs more than a detection flag, they need fault diagnosis. This is what distinguishes COMET from a single OOD detector. The ablation shows that the strongest single monitor (S1c) can only detect that something is off, whereas COMET detects and isolates faults, coordinate the outputs of the different monitors and communicates it by this separates a dead sensor from an implausible detection from an unfamiliar context, each implying a different supervisor action. This aligns with the fault-tolerant system design of \citep{DiagnosisFaultTolerantControl2006}, and considers a human supervisor and its cognition.

However, this principle is not scalable by simply adding more monitors to cover every conceivable fault. Such an extension may be useful for offline analysis, but online it can produce an information overflow that achieves the opposite of an informed supervisor if not adequately combined with coordination and communication layers. The right monitor selection therefore depends on the system setup like how fast decisions must be made under production pressure, and what the control interface can meaningfully display. In COMET, this decision can be made explicit twice, i.e., in deciding the monitor structure, and in designing the coordination and communication layers that combine the decentralized solution from the different monitors and compress the resulting taxonomy into a traffic-light-based communication. If human supervisors carry the final responsibility, they must be enabled to take informed decisions, and COMET aims to fulfill this requirement.

At the same time, diagnostic richness has its own risk. COMET's outputs on what is wrong may also mislead. The taxonomy states are hypotheses, not ground truth, and a wrong attribution can push the human supervisor into a diagnostic direction that is itself incorrect. Our results contain concrete instances where microphone-3's deviating template produced spurious consistent verdicts that would have presented a novel fault as trusted leak physics, and on tube leaks a small share of hidden faults was surfaced as T4 (unfamiliar context) rather than T2 (hidden fault). It is correct to escalate, but pointing at the context rather than the fault. The interface should therefore present taxonomy states as starting points for investigation and preserve access to the underlying outputs of each monitor, rather than as settled diagnoses. And the residual silent misses show the complementary failure, i.e., absence of an alarm is not evidence of safety. Training supervisors on both fault directions is as much part of the monitoring system as the individual monitors themselves.

\subsection{Signal-integrity monitor}
The signal-integrity monitor's strengths are practical where its rules are understandable and adjustable by engineers, and it is cheap enough to run in real time on any hardware. It is also deterministic, so for the faults within its rule envelope it provides a guarantee rather than a statistical statement. In our experiments, dropout, flat-line, and offset injections were intercepted at 100\%. For such well-specified sensor faults, fixed rules are the right tool, and a learned OOD detector would be an overpowered and less transparent substitute.

At the same time, our results expose the limits of this monitor. The clipping rule failed in both directions. It detected only 9.6\% of soft-clipping injections, and it over-fired on microphone-1 (15\% of nominal no-leak files). Both defects are specification-level, not learned, and therefore fixable by rule revision. A revision is a specification change that must be re-validated on nominal data under the same freeze discipline as every other component, not an ad-hoc adjustment at runtime. The general lesson is that hand-crafted rules are auditable, but auditability is not correctness. Integrity rules need the same validation discipline as the learned components they supervise. A deterministic guarantee covers exactly the faults that were specified which is why this layer complements, and cannot replace, the physics and feature-space layers.

\subsection{Physics-consistency monitor}
The physics-consistency monitor illustrates both why purely model-based leak detection is difficult and why physical knowledge remains valuable as a complement. Its behavior also points to a caution for physics-informed machine learning in general. Embedding physics into a learned model via loss function promotes physically plausible behavior but does not necessarily provide guarantees \citep{bruntonOptimizationBootcampMachine2026}, and once the environment shifts such that the underlying physical assumptions no longer hold, the physics baked into the model become a silent fault itself. It is entangled in the learned weights and invisible from the output. Our design avoids this entanglement by keeping the physics outside the model, as a separate, auditable component. The microphone-3 template anomaly shows why this matters. The deviating template fit was a degraded physical assumption, and it was detectable precisely because the monitor's verdicts are explicit and comparable across microphones, rather than absorbed into a network.

The results show the monitor's characteristic behavior. It is often silent, abstaining on 83--88\% of files under covariate shift, but silence is not indecision. It is the honest answer when the background dominates the leak bands, and it degrades to no judgment rather than to a wrong one. When the monitor flags, its statements were reliable. Zero false inconsistencies on nominal leaks, and zero physics-supported false alarms under covariate shift. This asymmetry is what makes it valuable as a confirmation layer whose silence and inconsistency both carry information. The physics-consistency monitor's template covers known leaks, so it can confirm but not falsify. An inconsistent verdict marks the boundary of known physics, not the boundary of leaks.

We stress that our implementation is deliberately simple. A fixed spectral template with an abstention gate. A more dedicated physical model, e.g., with flow-rate or geometry dependence, could be more sensitive and robust, but the present monitor suffices to demonstrate the concept. The larger scope limitation is that not every task offers physics that can be turned into a monitor as directly as jet acoustics. Where such knowledge exists, however, our results show that using it as an independent consistency check, rather than as the detector, or as a constraint inside the detector, is a useful way of bringing physical knowledge into AI-enabled safety systems.
 
\subsection{Feature-space monitor}
The feature-space monitor is, at its core, OOD detection, and it was the most powerful single monitor in our experiments. Its results are consistent with the broader literature that OOD detection works, but imperfectly, as shown both in the original k-NN study \citep{sunOutofDistributionDetectionDeep2022} and in recent surveys \citep{yangGeneralizedOutofDistributionDetection2024}. The underlying challenge is the clean separation of ID and OOD distributions by a scalar score, i.e., in our case the k-NN distance. This separation is hard to achieve at all in high-dimensional spaces \citep{fangOutofDistributionDetectionLearnable2022}, and even a separation that holds on training and test data is not guaranteed to persist at runtime, which is why periodic recalibration of the monitor remains an open operational question. In our results, the imperfect separation is directly visible where part of the shifted data lies too close to the nominal reference in feature space to be flagged. Most notably the residual tube-leak files that passed all layers (Section~4.6.4).

A structural reason became apparent during development where the penultimate space is heavily collapsed, with a participation ratio of 1.4--2.5 out of 64 dimensions. The binary task does not require more, and the monitor still works, but the space is far from a rich representation. We attempted a supervised-contrastive retraining to spread the embedding. It failed its pre-registered acceptance criterion, so the cross-entropy encoder stayed frozen. \citet{fangOutofDistributionDetectionLearnable2022} suggest that training with more classes can improve OOD separability. We did not pursue this, since our AI system deliberately represents a traditional safety-related classifier, which is typically binary. Testing whether multi-class training yields a better embedding for monitoring is future work.

Two further observations qualify the monitor's role. First, as a single monitor it is fast, but its information content is low. It reports only that something is unfamiliar, not what or why. The diagnostic value emerges only in combination with the other monitors. Second, the layer-profile variant showed that there is more structure to extract. Its per-layer z-profiles separated covariate from semantic shift and provided genuine diagnostics. Its elevated false-alarm rate on nominal leak files (34.4\%) disqualified it as the L3 main vote, but further investigation may turn it into a robust monitor in its own right.

\subsection{Coordination and communication layer}

The coordination layer and its taxonomy's purpose is to combine the monitors' outputs into states with different urgency. For a human supervisor to understand and act on quickly, the communication layer is implemented. The mapping of taxonomy states to a traffic-light-based display with action recommendations is a part of a deliberate trade-off. Compressing monitors' individual outputs into a small set of states with a traffic-light urgency mapping enables fast decisions, but any such compression also pre-structures the supervisor's judgment. On the other hand, a raw display of monitor outputs would leave more room for the operator's own assessment of whether a situation is critical but quickly leads to an information overflow. How much abstraction is right, and how the communication layer should be designed, cannot be answered by system developers alone. It must be co-designed with the operators who will use the system, in their actual decision context. The coordination layer, its taxonomy, and the communication layer, its traffic-light-based design, proposed in this work should therefore be read as one proposal of how such a concept can look and not as the final design.

The coordination layer also defines the framework's extensibility. In the present work it is deliberately simple: a static, rule-based taxonomy derived from engineering intuition and frozen before evaluation. The architecture allows this layer to grow in at least three directions. First, further monitors can be stacked onto the coordination layer without changing its logic, since the taxonomy consumes verdicts rather than monitor internals. Second, the coordination could incorporate beliefs about the system state and how new observations update them. At deployment, the implicit belief is that the data resembles the training distribution. Over time, sampling becomes biased rather than i.i.d., even if inputs remain inside the competence envelope, and this gradual shift affects performance and should inform the coordination. Third, prior reliability knowledge can be integrated. If a sensor's failure rate as a function of lifetime is known, e.g., expected failure after ten years, this prior can modulate how much weight its monitor's verdicts receive, turning the static taxonomy into a dynamically updated, and thus "smarter", coordination layer. We note that such dynamic updating would need its own governance. A coordination layer that changes at runtime must itself be validated, extending the freeze-before-evaluation discipline applied in this work.

\subsection{Ablation}
The ablation ladder most clearly underlines the need for runtime monitoring of AI systems. Without any monitor, every AI system misclassification is silently trusted (100\% silent-miss share, TWA up to 58.3\%). Physics alone (S1b) improves this only marginally and it abstains exactly where the AI system fails, leaving 96.6\% of missed tube leaks silent. MSP, a popular OOD baseline, improves the silent-miss share substantially over no monitoring (63.2\% on tube leaks) and moderately reduces TWA, but remains insufficient as a safeguard. The k-NN monitor (S1c) and COMET (S2) are the two strongest options. On tube leaks, TWA is 8.5\% vs.\ 8.3\% and silent-miss shares are 28.2\% vs.\ 27.6\%. The COMET's quantitative advantage appears mainly under covariate shift, where the integrity and physics layers eliminate the 10.1\% trusted-wrong answers that S1c leaves on hydraulic no-leak data, and on injected faults, which S2 intercepts with attribution. The price is a small nominal false-alarm cost.

Honestly, we expected COMET to outperform the single monitors more clearly than it did. We report these findings as they are, because all calibrations were frozen after development and the shifts were evaluated zero-shot. The comparison was not tuned in COMET's favor. One legitimate reading of the results is that a single k-NN monitor is already a powerful and cheap heuristic for detection. We would still argue for the COMET framework. It is quantitatively equal or better on every condition in terms of fault detection, but it detects and isolates fault types that a feature-space monitor cannot attribute. Thus, its outputs are informative in the moment, i.e., separating sensor faults from implausible detections from unfamiliar context. It also supports offline analysis and maintenance afterwards. In safety-critical operation, that diagnostic content, not the marginal TWA difference, is the deciding factor for us.

\subsection{Limitations}
First, this study evaluated the framework on a single dataset, and extension to further datasets and applications is necessary future work. The practical challenge is finding data that represents, or can serve as a proxy for safety systems and contain shifts that qualify as covariate and semantic. The shifts must also be realistic. Synthetic shifts can be overly constructed, making them both unrealistic and potentially too easy to detect for a monitoring system.

Second, the physics-consistency layer presumes that the task offers physics that can be modeled. This is not always the case. Where it is not, COMET can still be pursued with a differently coordinated structure of monitors. The central concept is not the specific three layers but the coordination of monitors with disjoint fault modes and the communication layer, which is what enables diagnosis rather than bare detection flags. How well the coordinated multi-layer principle generalizes to other domains should be tested in future work.

Third, at the system level, monitoring an AI system does not make the system automatically safe. Monitoring increases the reliability of the AI system and can have positive implications for system safety, but the overall system is a socio-technical one, and safety must be addressed at that level with procedures, fallbacks, and organizational measures that lie outside this work. An interesting direction is to investigate how integrating a runtime monitoring framework such as COMET affects system-level safety.

Finally, our claim that COMET is more informative for operators than a single monitor is currently supported by the architecture, the ablation, and fault-tolerant system theory, not by evidence with human operators. Validating it requires field studies, ideally conducted together with human-factors and social-science experts both to test whether supervisors actually make better and faster decisions with the coordinated structure, and to co-design the monitors and their presentation with the people who will rely on them.

\section{Conclusion}
This work presented COMET, a coordinated multi-layer runtime monitoring framework for human-supervised AI-enabled safety systems. In contrast to the single monitors typically studied in research, COMET combines monitoring layers that target different fault modes and thereby enable fault diagnosis by the human supervisor in the loop. A coordination layer obtains the outputs of different monitors and combines them into taxonomy states. Then, these states are communicated to a human supervisor that acts as adaptive controller for the system. In this work, the framework consists of three monitors: a signal-integrity layer that detects typical sensor faults such as flat-line or missing data, a physics-consistency layer that checks whether detections match a spectral leak template derived from the physics observed in nominal training data, and a feature-space layer that assesses whether an input lies inside the AI system's learned competence envelope. The layer outputs are obtained and combined into taxonomy states by the coordination layer and then present via traffic-light-based display to the supervisor, supporting fast, intuitive decisions at runtime.

The framework was evaluated on an open compressed-air leak detection dataset, a proxy for safety-critical gas leak detection, with an AI system classifying inputs as "leak" or "no leak". Held-out background noises and a novel leak type provided realistic covariate and semantic shifts, complemented by injected sensor faults. All monitors were calibrated on nominal data only and frozen before a single zero-shot evaluation of the shifts. Without monitoring, the shifts produced confident, silent misclassifications and environment changes that remained invisible to operators. COMET detected, isolated and correctly identified most of them. It reduced the trusted-wrong-answer rate to zero on covariate-shifted no-leak data and on three of four injected sensor faults, at a nominal false-alarm cost of $\approx 1\%$. Compared to a single OOD detector which detected shifts but could not attribute them COMET added fault isolation and fault identification in some cases without sacrificing detection performance. At the same time, the evaluation delineated the boundary of passive monitoring with 27.6\% of the AI system's undetected tube leaks passed all monitors silently, a residual risk that further layering of passive monitors cannot be expected to remove.

Overall, this work demonstrated that coordinating multiple monitors with disjoint failure modes is an effective way to increase the reliability of AI systems, which is crucial in safety-critical systems and other high-risk applications. Beyond detection, the layered structure supports fault diagnosis by isolating and in some cases also identifying faults, thus informs human decision-making at runtime, and targets updating and maintenance afterwards. Future work should extend the evaluation to further datasets and applications, validate the coordination layer and its taxonomy with human operators, and investigate how such monitoring contributes to safety at the socio-technical system level.

\section*{Acknowledgments}
This work was carried out as a part of SUBPRO-Zero, a Research Center at the Norwegian University of Science and Technology (project code 100589 and CristinID 2720670). The authors gratefully acknowledge the project support from SUBPRO-Zero, which is financed by  major industry partners and NTNU.


\printcredits

\section*{Declaration of generative AI and AI-assisted technologies in the manuscript preparation process}

During the preparation of this work the author used Kimi Code to assist in building the code base and running tests. After using this tool, the author reviewed and edited the content as needed and takes full responsibility for the content of the published article.

\bibliographystyle{cas-model2-names}

\bibliography{cas-refs}

\end{document}